\pdfoutput=1

\documentclass[11pt]{article}
\usepackage[utf8]{inputenc}

\usepackage{acl}
\usepackage{graphicx}
\usepackage{colortbl}
\usepackage{times}
\usepackage{latexsym}
\usepackage{multirow}
\usepackage{booktabs}
\usepackage{siunitx}
\usepackage{caption}
\usepackage{enumitem}

\usepackage{amsmath}

\usepackage[T1]{fontenc}

\usepackage[utf8]{inputenc}
\usepackage{listings}
\usepackage{xcolor}

\lstdefinelanguage{json}{
    morestring=[b]",
    morecomment=[l]{//},
    morekeywords={true,false,null},
    sensitive=false,
}

\usepackage{inconsolata}
\newcommand{\getreason}{\textbf{\textsc{GETReason}}}
\newcommand{\great}{\textsc{GREAT}}

\title{GETReason: Enhancing Image Context Extraction through Hierarchical Multi-Agent Reasoning}

\author{
 Shikhhar Siingh$^{*}$\quad
    Abhinav Rawat$^{*}$\quad
  Chitta Baral \quad
  Vivek Gupta$^{\dag}$ \\
  School of Computing and Augmented Intelligence, Arizona State University \\
  \texttt{\{ssiingh, arawat21, chitta, vgupt140\}@asu.edu}
}

\begin{document}
\maketitle
\def\thefootnote{*}\footnotetext{These authors contributed equally to this work.}
\def\thefootnote{\dag}\footnotetext{This corresponding author supervised the research.}
\begin{abstract}
Publicly significant images from events carry valuable contextual information with applications in domains such as journalism and education. However, existing methodologies often struggle to accurately extract this contextual relevance from images. To address this challenge, we introduce \getreason{} (\textbf{G}eospatial \textbf{E}vent \textbf{T}emporal \textbf{Reason}ing), a framework designed to go beyond surface-level image descriptions and infer deeper contextual meaning. We hypothesize that extracting global event, temporal, and geospatial information from an image enables a more accurate understanding of its contextual significance. We also introduce a new metric \great{} (\textbf{G}eospatial, \textbf{R}easoning and \textbf{E}vent \textbf{A}ccuracy with \textbf{T}emporal alignment) for a reasoning capturing evaluation. Our layered multi-agentic approach, evaluated using a reasoning-weighted metric, demonstrates that meaningful information can be inferred from images, allowing them to be effectively linked to their corresponding events and broader contextual background.
\end{abstract}

\section{Introduction} 

Public event images capture moments of societal and historical importance—presidential inaugurations, mass protests, international summits—serving not only as visual records but as rich contextual artifacts. Understanding such images requires more than surface-level description; it demands inference about geopolitical, temporal, and event-specific factors often implicit in the scene. Despite their significance for journalism, archival analysis, and public discourse, event-centric image understanding remains underexplored.

\begin{figure}[h]
    \includegraphics[width=3in]{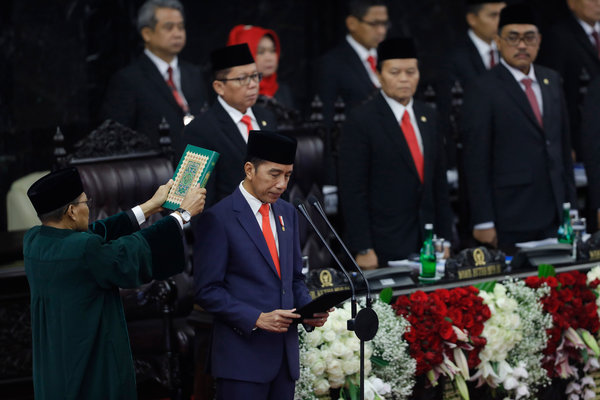}
    {\small
    \noindent \textit{location:} Indonesia ; \textit{time:} October 20, 2019 \\
    \textit{event:} Indonesian President Joko Widodo’s 2$_{\text{nd}}$ term Inauguration.}
    \vspace{-0.5em}
    \caption{\small Example from TARA~\citet{FZCVR22} dataset. Given an image, extract location and time. We add the inferred event.}
    \label{fig:indonesian_president}
\vspace{-1.2em}
\end{figure}

Most existing vision-language methods fall short in this setting. Traditional captioning models and encoder-decoder architectures describe visible content—objects, people, and actions—but fail to infer deeper meaning. Even advanced visual-language models~\cite{li2022mplug, hu2023exploiting} tend to narrate what is seen, missing why it matters. For instance, given an image of President Joko Widodo’s inauguration (Figure \ref{fig:indonesian_president}), these models might mention key figures but omit the event’s political gravity and temporal context. However, these methods fall short of true event understanding, as they fail to reason about temporal, spatial, and sociopolitical factors—many of which are implicit rather than directly observable.



Efforts to move beyond superficial cues include reasoning-based methods~\cite{zhang2024goodguesser, song2024cogbench}, which can infer coarse event types like “ceremony,” and retrieval-augmented generation (RAG)\cite{lewis2020retrieval}, which brings in external knowledge. Yet both have limitations: the former lacks specificity, and the latter is vulnerable to hallucinations and misinformation due to noisy sources\cite{zeng2025worse, deng2024cram}. These approaches often miss the nuanced, high-precision understanding required for real-world applications. Therefore, a robust solution is needed to automatically generate richer and more accurate narratives of public events, enhancing both human comprehension and downstream machine applications.


Recent advances in multi-agent frameworks~\cite{dinh2025entagents,ghafarollahi2024sciagents,wang2025jupybara,ng2024pseudo} demonstrate that dividing reasoning tasks among specialized agents can significantly improve contextual understanding across domains. However, these methods have primarily focused on textual or code-based settings. This raises an important question: \textit{Can a structured multi-agent framework effectively support contextual reasoning over complex visual data, such as public event images?}

To answer this question, we introduce \textbf{\getreason{}}, a structured multi-agent framework for extracting rich, contextual narratives from public event images. \getreason{} decomposes the task into three core sub-problems, each handled by a dedicated agent: (1) geospatial inference (e.g., “Indonesia”), (2) temporal inference (e.g., “October 20, 2019”), and (3) event-specific interpretation (e.g., “inauguration of President Joko Widodo’s second term amid political unrest”). Building on the divide-and-conquer principles of recent multi-agent systems~\cite{ng2024pseudo,dinh2025entagents}, \getreason{} enables collaborative cross-agent validation to enhance factual accuracy, reduce hallucinations, and improve generalization across diverse event types.

We evaluate \getreason{} on a large-scale dataset of public event images and demonstrate substantial gains over existing captioning and reasoning baselines. Our results highlight the importance of structured, role-specialized reasoning in achieving precise and reliable event understanding—unlocking new capabilities for downstream applications in news generation, historical archiving, and multimodal analysis. In summary, our key contributions are as follows:

\begin{itemize}
    \vspace{-0.75em}
    \item \textbf{Structured Multi-Agent Extraction Framework:} We introduce a framework that systematically extracts event-related details while mitigating hallucinations~\cite{xu2018survey}, ensuring that VLMs focus on accurate and contextually relevant information rather than generating misleading inferences.
    \item \textbf{Evaluation Metric for Reasoning Quality:} We propose a novel yet straightforward evaluation metric that inherently incorporates the reasoning process, allowing for a more meaningful assessment of the quality of extracted information.
    \vspace{-0.75em}
    \item \textbf{Augmentation of Event-Centric Datasets:} We address limitations in existing datasets by augmenting them with the attributes required for effective event extraction, ensuring better alignment with the problem domain.
    \vspace{-0.75em}
    \item \textbf{Systematic Evaluation of Reasoning Strategies:} Using carefully selected subsets of the augmented datasets, we rigorously evaluate the reasoning capabilities of VLMs, analyzing the specific strategies they employ to extract event-related details.
\end{itemize}

The code and dataset are available at \url{https://coral-lab-asu.github.io/getreason/}

\section{Task Description}

We define the task of \textbf{event-centric image understanding} as generating a structured triplet from a single image depicting a public event. Given an image $I$, the model must infer: (1) a \textbf{location} $L$ (e.g., city or country), (2) a \textbf{temporal reference} $T$ (e.g., specific date or period), and (3) an \textbf{event description} $E$ that captures the scene’s sociopolitical significance. Unlike standard captioning, which describes visible entities or actions, this task requires reasoning about partially implicit geopolitical, temporal, and narrative cues.

In Figure~\ref{fig:indonesian_president}, for instance, the model links the image of President Joko Widodo to an oath-taking ceremony, the green-bound 1945 Constitution, and the presence of key political figures to derive the structured output: \textit{Expected Output}: $L$: ``Jakarta, Indonesia''; $T$: ``October 20, 2019''; $E$: ``Inauguration of President Joko Widodo’s second term amid protests.''

Each component involves distinct reasoning challenges: $L$ depends on geographic signals (e.g., flags, landmarks); $T$ draws on visible dates, seasonal features, or known timelines; and $E$ requires synthesizing the visual context with latent knowledge of political and historical events. This formulation enables structured, interpretable outputs for applications such as news analysis, archival indexing, and societal trend monitoring, while posing challenges in multimodal inference and hallucination control.

\section{Dataset Description}

We use two dataset for our evaluation: WikiTilo~\cite{zhang2024goodguesser} and TARA~\cite{FZCVR22} dataset. These dataset have images with temporal and geospatial information. 

\paragraph{WikiTiLo Dataset.}

The WikiTiLo dataset consists of 6,296 images annotated with specific spatio-temporal information, covering over 30 countries and spanning the years from 1826 to 2021. This dataset provides a diverse range of images with annotations detailing the time and country where each image was captured.

\paragraph{TARA Dataset.}
The TARA dataset comprises of approximately 16,000 images associated with news articles, along with their corresponding time and location information, automatically extracted from the New York Times (NYT)\textsuperscript{}\footnote{\url{https://developer.nytimes.com/docs/archive-product/1/overview}}. For our evaluation, we utilized the train set which comprises of 12,306 samples from this dataset, spanning from January 2010 to May 2021. Each image is accompanied by metadata detailing the time and location pertinent to the associated news article.

\vspace{0.5em}
\noindent \textit{Data Sampling.} Out of the 12,306 samples from TARA, we excluded 1,065 images due to incorrect formats and socially inappropriate content as shown in Table \ref{tab:dataset_statistics}. As a result, 11,241 images from TARA were retained for our study. In contrast, we can see in Table \ref{tab:dataset_statistics} that the entire WikiTiLo dataset of 6,296 images was utilized without exclusions. However, due to the absence of accompanying textual context, as found in the TARA dataset, certain components of our analysis were adjusted accordingly as discussed in section \ref{sec:tara_star}.

\begin{table}[h]
\small
\centering
    \begin{tabular}{lccc}
        \toprule
        Dataset & Total  & Excluded & Utilized  \\
        \midrule
        TARA     & 12,306 & 1,065 & 11,241 \\
        WikiTiLo & 6,296  & 0     & 6,296  \\
        \bottomrule
    \end{tabular}
    \vspace{-0.5em}
    \caption{ \small Summary of datasets images utilized in the study.}
    \label{tab:dataset_statistics}
    \vspace{-1.5em}
\end{table}

\subsection{TARA*: TARA Enhancement}
\label{sec:tara_star}

To better suit our task, we augment TARA dataset with event and fine-grained temporal and geospatial information generated from the metadata present in the ground truth using a Vision Language Model (VLM), specifically, Gemini 1.5 Pro\textsuperscript{}\footnote{\url{https://deepmind.google/}}.

\paragraph{Spatio-Temporal Augmentation.}
The metadata of the TARA dataset provided time and location details. This metadata was processed separately using the VLM to structure the information into a predefined JSON format, categorizing temporal data into century, decade, year, month, and day, and spatial data into country, state/province, and city. 

\paragraph{Event Information Augmentation.}

The image, along with its associated news article, was provided to the VLM to simultaneously generate the primary event, the secondary event serving as background context, and the corresponding reasoning for how these events are grounded in the image. The resulting information was then stored in a predefined JSON format.

The outputs from these two phases were combined to form the final ground truth labels, providing comprehensive event, spatial, and temporal context for each image used in the evaluation.

\paragraph{Deduction Augmentation.} Finally, we used the augmented ground truths as references for the images on the VLM to generate the deduction inferences and reasoning from the images for all the information present in the augmented dataset. For example, in Figure \ref{fig:indonesian_president}, the event deduction of the image will be:
The image shows Joko Widodo taking an oath on the Quran during his initiation ceremony, the temporal deduction of the image will be: the image is about the initiation ceremony of Joko Widodo which took place in 20 October, 2019 and the geospatial ceremony of the image will be: from the attire of the people in the background and the setting, it looks like a government setting in Jakarta. Since the event is inauguration of Joko Widodo, the geospatial location is at the People's Consultative Assembly (MPR) Building. 

\subsection{WikiTiLo*: WikiTiLo Enhancement} In WikiTiLo, there was no associated meta-data hence, the event cannot be extracted. However, we restructured the ground truth labels to suit our task. The existing annotations were restructured to align with the format used for the TARA dataset, facilitating a consistent  evaluation framework.

\section{{\sc GETReason} Architecture}

\begin{figure*}[t]
    \centering
    \includegraphics[width=0.98\textwidth]{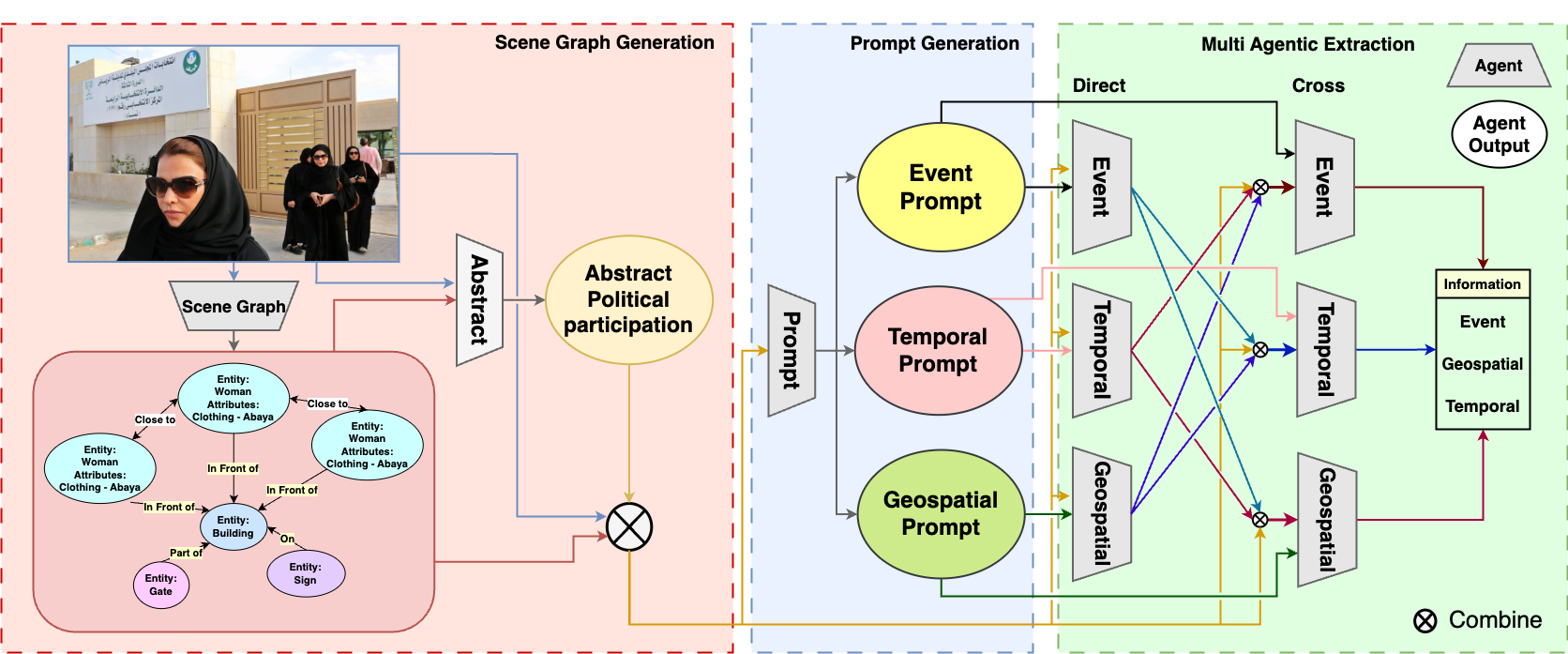}
    
    \caption{ \small GETReason architecture}
    \label{fig:GETReason_architecture}
\vspace{-1.25em}
\end{figure*}

Our framework comprises of three layers: Scene Graph Generation, Prompt Generation, and Multi-Agentic Extraction. Each layer consists of VLM agents that generate outputs based on specific prompts, working collaboratively to produce comprehensive and contextually rich captions.

\subsection{Scene Graph Generation}

The Scene Graph Generation serves as the primary process of our framework, responsible for extracting and structuring information from the input image. It comprises two main modules:

\paragraph{Scene Graph Agent.}

This module identifies entities within the image, along with their attributes and the relationships between them. By analyzing the visual content, it constructs an open-ended, yet structured representation that captures the spatial and semantic interactions present. For instance, in an image depicting women participating in voting for the first time, the generator identifies three women, notes their clothing attributes, and captures relationships such as the women standing in front of a building and being close to each other as can be seen in the Figure \ref{fig:GETReason_architecture}. This structured output is formatted as a JSON object, detailing the entities, their attributes, relationships, and the reasoning behind these relationships.

\paragraph{Abstract Agent.}

Building upon the initial scene graph, this module enhances it by inferring the broader abstract concept conveyed by the image. This involves analyzing both the visual elements and the initial scene graph to deduce higher-level interpretations. Continuing from the previous example, analyzing the image and its scene graph might lead to the inference that the abstract idea pertains to women's participation in the Saudi Arabian political process, based on cues like traditional clothing and signage indicating a polling station in Riyadh. This enriched scene graph provides a more comprehensive understanding, which can help in generating more contextually relevant responses in subsequent processes.

\subsection{Prompt Generation}

The Prompt Generation consists of the Prompt Agent which is designed to generate tailored prompts for the agents in the Multi-Agentic Extract, ensuring precise and context-specific analyses. By creating context, role, and image-specific prompts, this agent directs each agent to focus on pertinent aspects of the image, enhancing the precision and robustness of their analyses. For instance, the prompt for the Geospatial Information Analyzer might direct attention to signage and clothing attributes to deduce the photo's location. This targeted prompting mechanism ensures that each agent operates within its domain expertise, leading to more accurate and relevant outputs.

\subsection{Multi-Agentic Extraction}

In this process, specialized agents utilize the contextual information represented by the scene graph and prompts to perform their designated analyses. They generate the "value" for their task as well as their deductional reasoning of arriving at their answer. This process consists of three agents:

\paragraph{Event Agent}

The Event Agent is designed to infer the primary event depicted within an image, extending beyond a surface-level summary of observable actions. Instead, it synthesizes information from the structured scene graph, an abstract concept generated in preceding processing stages, and relevant world knowledge. According to the methodology described in the Appendix, the Event Agent’s prompt directs it to articulate detailed reasoning grounded in structured contextual cues, and make logical and well-supported assumptions in instances where direct evidence is absent, thereby ensuring both relevance and factual accuracy. For example, the agent may integrate signals such as traditional attire, public signage, and the presence of recognizable figures to infer the occurrence of a political or cultural event, elaborating on the rationale behind such inferences. The structured output of the Event Agent is a JSON object containing fields for the event, event reasoning, background, and background reasoning, which ensures consistency and minimizes hallucinations

\paragraph{Temporal Agent}

The Temporal Agent is tasked with extracting fine-grained temporal information from the depicted scene, including the century, decade, year, month, and day. Its operation, as mentioned in the Appendix, is guided by a prompt that encourages the use of diverse cues—such as lighting conditions, the presence of celestial objects, and indications of technological or architectural styles—as well as textual information (e.g., signage) available within the image or the corresponding scene graph. The agent is required to justify its reasoning at each level of temporal granularity, or, where direct evidence is incomplete, to provide the most plausible estimate grounded in available context. The Temporal Agent produces a structured JSON output containing the century, decade, year, month, day, and accompanying reasoning, which enhances reliability and mitigates error propagation.

\paragraph{Geospatial Agent}

The Geospatial Agent is specialized for precise image localization, outputting structured geospatial information at the levels of country, state or province, and city. As outlined in the Appendix, the agent’s prompt emphasizes a comprehensive assessment of available visual cues—including, but not limited to, signage, clothing, architectural features, and explicit geotags—supplemented by the abstract scene description. The agent is required to rigorously justify each localization decision and to refrain from speculation in the absence of sufficient evidence. Its structured output is a JSON object detailing the country, state or province, city, and the agent's reasoning, which supports robust and interpretable geospatial predictions.

\paragraph{Direct Extraction}
We first employ a direct extraction from each of our three agents to extract temporal, event, and geo-spatial information. Each agent processes the image, the scene graph enhanced with abstract idea, and the corresponding prompt to produce its analysis, contributing to a holistic understanding of the image's context.

\paragraph{Cross Extraction}

Multi-Agentic Extraction offers valuable insights but may introduce hallucinations due to the agents' reliance on best-guess predictions and incomplete information. Additionally, the extraction process may be less efficient without the influence of other agents' responses to refine outputs. To enhance accuracy, we implement a two-stage iterative reasoning strategy through cross-extraction. This involves feeding previously gathered contextual clues from all other agents back into each agent, enabling deeper reasoning and reducing hallucinations. By iteratively refining their analyses with a richer context, the agents can achieve more comprehensive and accurate interpretations.

In summary, our methodology integrates structured scene representation, targeted prompting, specialized agent analysis, and iterative reasoning to generate detailed and contextually enriched image captions.

\section{\great{} Evaluation Metric.}
Various metrics have been used to evaluate reasoning in LLM-based techniques, each with its strengths and limitations. Standard metrics like accuracy, F1-score, precision, and Recall @ K rely on exact matching, making them unsuitable for tasks requiring semantic understanding ~\citet{powers2011evaluation}.  

More advanced metrics such as ROUGE ~\citet{lin2004rouge}, BLEU ~\citet{papineni2002bleu}, and METEOR ~\citet{banerjee2005meteor} improve evaluation but remain biased toward word overlap and order, limiting their ability to capture meaning. ROUGE also favors longer texts ~\citet{lin2004rouge}, while CIDEr ~\citet{vedantam2015cider}, though effective for image captioning, is not designed for reasoning tasks requiring spatio-temporal evaluation. Human evaluation, while useful for qualitative assessment, lacks quantitative consistency.

For geospatial and temporal evaluation, a simple matching based F1 scoring can be used like in~\cite{FZCVR22}, however, a major drawback of this metric is that it does not score based on the reasoning ability of the model, i.e. the closeness of the value to the ground truth is not measured. Thus, to address these issues, we propose a new metric - \great{}, which comprises of three parts: 

\paragraph{Event Evaluation.}

In \great{}, the event score is computed using cosine similarity (CS) between the embeddings of the concatenated event and background values for both the predicted and ground truth outputs. This approach leverages the embeddings generated by a Sentence Transformer model to measure semantic similarity effectively.

Let \( E_i \) and \( B_i \) denote the augmented event and background for image \( i \) and let \( e_i \) and \( b_i \) denote the corresponding predicted event and background values. The concatenated event and background representations are given by \( e_i + b_i \) for predictions and \( E_i + B_i \) for ground truth labels. We first compute the cosine similarity between the embeddings generated by a Sentence Transformer model from HuggingFace~\citet{wolf2019huggingface}, specifically the \texttt{all-mpnet-base-v2} model, for the predicted and ground truth concatenated event-background pairs. Since the cosine similarity lies within the range \([-1, 1]\), we shift this range to \([0, 1]\) by using the formula:

\vspace{-0.5em}
{\small
\[
\text{CS}_{\text{shifted}} = \frac{\text{CS} + 1}{2}
\]
}
\vspace{-0.5em}
\noindent Thus, the event score for image \( i \), (\( \text{ES}_i \)) is defined:

{\small
\[
\text{ES}_i = \frac{\text{CS}((\text{e}_i + \text{b}_i), (\text{E}_i + \text{B}_i)) + 1}{2}
\]
}
\vspace{-0.0em}
\noindent By leveraging cosine similarity, \great{} measures the semantic similarity between the predicted and ground truth event-background pairs, effectively evaluating model’s reasoning strength.

\paragraph{Geospatial Evaluation.}

To assess the accuracy of predicted geospatial locations, we define a similarity metric that evaluates the hierarchical correctness of location predictions and computes a distance-based score using the Haversine formula. Given a ground truth location \( gt \) and a predicted location \( pred \), we first determine the deepest available location level among \textit{country}, \textit{state/province}, and \textit{city}. If no valid prediction exists, the score is set to 0. The geospatial similarity is then computed based on the great-circle (Haversine) distance~\citet{haversine} between the latitude-longitude coordinates of the ground truth and predicted locations. The Haversine distance \( d \) is given by:  

\vspace{-1.25em}
\begin{equation}
\scalebox{0.7}{$d = 2 R \arcsin \left( \sqrt{\sin^2 \left( \frac{\Delta \phi}{2} \right) + \cos(\phi_1) \cos(\phi_2) \sin^2 \left( \frac{\Delta \lambda}{2} \right) } \right)$}
\vspace{-0.5em}
\end{equation}  

where \( R \) is the Earth's radius (typically \( 6371 \) km), \( \phi_1 \) and \( \phi_2 \) are the latitudes (in radians), and \( \lambda_1 \) and \( \lambda_2 \) are the longitudes (in radians). The terms \( \Delta \phi \) and \( \Delta \lambda \) represent the differences in latitude and longitude, respectively, between the ground truth and predicted locations:  

\vspace{-0.5em}
{\small
\begin{equation}
\Delta \phi = \phi_2 - \phi_1, \quad \Delta \lambda = \lambda_2 - \lambda_1
\end{equation}
}

\vspace{-0.5em}
\noindent The final geospatial similarity score is computed:

\vspace{-0.5em}
{\small
\begin{equation}
    S_{\text{geo}} = \max \left( 0, 1 - \frac{d}{D_{\max}} \right)
\end{equation}
}

\vspace{-0.5em}
where $D_{\max}$ is a predefined maximum distance threshold (default: $1000$ km). This formulation rewards predictions that are closer to the ground truth with higher scores, while penalizing those that are farther away. Therefore, the better a model can accurately reason about the geospatial location, the closer its predictions will be to the ground truth, resulting in a higher score. Additionally, if all location fields in the ground truth are missing, the score defaults to $1$ to reflect the irrelevance of geospatial constraints in such cases.

\begin{table*}[t]
\small
\centering
\setlength{\tabcolsep}{4.5pt}
\begin{tabular}{l l cccc cccc cccc}
\toprule
 & & \multicolumn{4}{c}{\textbf{Gemini 1.5 Pro-002}} & \multicolumn{4}{c}{\textbf{QwenVL2.5-7B-Instruct}} & \multicolumn{4}{c}{\textbf{GPT-4o mini}} \\
\cmidrule(lr){3-6} \cmidrule(lr){7-10} \cmidrule(lr){11-14}
\textbf{Dataset} & \textbf{Method} & Geo & Temp & Event & Total & Geo & Temp & Event & Total & Geo & Temp & Event & Total \\
\midrule
\multirow{10}{*}{TARA} & COT$_{\text{zeroshot}}$ & 51.1 & \underline{37.7} & 66.5 & 53.3 & 29.7 & 8.2 & 61.2 & 35.9 & 39.2 & 36.8 & \underline{66.3} & 49.3 \\
& DA Prompt & 48.6 & 36.3 & \underline{67.6} & 52.5 & 36.6 & 9.8 & \underline{62} & 38.7 & 41.6 & 39.6 & 65.2 & 50.4 \\
& CogBench & 37.8 & 31.3 & 63.3 & 46.1 & 24.2 & 4.4 & 60.3 & 32.7 & 32 & 36.2 & 63.7 & 45.9 \\
& VIPHY & 32.1 & 31.9 & 62.3 & 44.1 & 32.2 & 5.7 & 61.0 & 35.8 & 28.2 & 36.7 & 63.8 & 45 \\
& EDIS & 32.7 & 32.2 & 63.5 & 44.8 & 32.8 & \underline{10.4} & 59.9 & 36.9 & 27.2 & 32.6 & 63.6 & 43.4 \\
& CAPTURE & 32.2 & 32.9 & 62.5 & 44.6 & 25.3 & 3.4 & 59.8 & 32.5 & 35.8 & 37.6 & 64.1 & 47.7 \\
& GTS & 47.8 & 34.2 & 65 & 50.6 & 37.7 & 8.1 & 61.1 & 38.2 & 39.8 & \underline{39.7} & 65.4 & 50 \\
& QR CLIP & 46.5 & 33.3 & 65.3 & 50.1 & 29.6 & 3.8 & 61.1 & 34.5 & \underline{45.2} & 39.4 & 66 & \underline{51.8} \\
& Good Guesser & \textbf{76.1} & 31 & 64.4 & \underline{57.8} & \underline{53.4} & \underline{10.4} & 56.8 & \underline{41.9} & \textbf{46} & 39 & 64.8 & 51.4 \\
    
\rowcolor{gray!10} & GETReason & \underline{69.4} & \textbf{38.1} & \textbf{70.3} & \textbf{60.4} & \textbf{60.5} & \textbf{23.3} & \textbf{65.3} & \textbf{51.3} & 45.1 & \textbf{41.9} & \textbf{68.5} & \textbf{53.5} \\
\midrule
\multirow{10}{*}{WikiTiLo} &  COT$_{\text{zeroshot}}$ & 26.7 & 27.1 & - & 26.9 & 23.9 & 18.1 & - & 21 & 14.4 & 25.7 & - & 20.1 \\
        & DA Prompt & 31.2 & 25.6 & - & 28.4 & 23.9 & 14.6 & - & 19.2 & 15.8 & 24.2 & - & 20 \\
        & CogBench & 32.1 & 25.8 & - & 29 & 25.1 & 12.8 & - & 18.9 & 12.5 & 21.8 & - & 17.2 \\
        & VIPHY & 12.2 & 19.3 & - & 15.7 & 18.3 & \underline{22.1} & - & 20.2 & 11.8 & 21.1 & - & 16.4 \\
        & EDIS & 19.2 & 26.3 & - & 22.8 & 19.4 & 14.2 & - & 16.8 & 10.9 & 19.4 & - & 15.2 \\
        & CAPTURE & 29.6 & 28.1 & - & 28.9 & 7.9 & 7.9 & - & 7.9 & 12.7 & 22.6 & - & 17.6 \\
        & GTS & 37.1 & 28.9 & - & 33 & 21.7 & 12 & - & 16.8 & 14.8 & 24.6 & - & 19.7 \\
        & QR CLIP & 37.4 & 27.7 & - & 32.5 & 27.8 & 11.6 & - & 19.7 & 15.9 & 24.9 & - & 20.4 \\
        & Good Guesser & \underline{40.2} & \underline{29.9} & - & \underline{35} & \textbf{38.5} & 17 & - & \underline{27.7} & \textbf{22.7} & \textbf{28.1} & - & \textbf{25.4} \\
        \rowcolor{gray!10} & GETReason & \textbf{42.4} & \textbf{34} & - & \textbf{38.2} & \underline{36.5} & \textbf{23} & - & \textbf{29.8} & \underline{17.6} & \underline
        {26.1} & - & \underline{21.9} \\
\bottomrule
\end{tabular}
\vspace{-0.75em}
\caption{ \small Evaluation Results(\%). \textbf{Abbreviations:} DA = Detective Agent, COT = Chain of Thought (\citet{wei2022chain})}
\label{tab:results}
\vspace{-2.0em}
\end{table*}

\paragraph{Temporal Evaluation.}
To assess the accuracy of the temporal predictions, we define a weighted scoring metric that evaluates predictions across multiple time granularities: century, decade, year, month, and day. Given a ground truth temporal annotation $gt$ and a predicted value $pred$, we compute individual scores for each time unit $u \in \{ \text{century, decade, year, month, day} \}$. The correctness of predictions is evaluated based on the absolute difference between $gt_u$ and $pred_u$, normalized according to the expected temporal variance at each level. Formally, the unit-wise score is defined as:
\vspace{-0.75em}
\begin{equation}
    \small
    S_u = 
    \begin{cases}
        1, \text{if } gt_u = pred_u \text{ (exact match, century level)} \\
        \max \left( 0, 1 - \frac{|gt_u - pred_u|}{T_u} \right), \text{~~~~~~~~~otherwise}
    \end{cases}
\end{equation}

\vspace{-0.5em}
where $T_u$ represents a unit-specific tolerance threshold: $T_{\text{decade}} = 50$, $T_{\text{year}} = 5$, $T_{\text{month}} = 6$, and $T_{\text{day}} = 15$. Each unit's contribution is further weighted based on its granularity, with higher precision levels (e.g., months, days) assigned greater importance. The final temporal accuracy score is computed as a weighted sum:

\vspace{-0.25em}
{\small
\begin{equation}
TS_i = \frac{\sum_{u} w_u S_u}{\sum_{u} w_u}
\end{equation}
}

\vspace{-0.5em}
where $w_u$ represents the predefined weight for each unit: $w_{\text{century}} = 1$, $w_{\text{decade}} = 1$, $w_{\text{year}} = 1.25$, $w_{\text{month}} = 1.5$, and $w_{\text{day}} = 1.5$. This formulation guarantees that predictions closer to the ground truth receive higher scores, while errors at finer granularities are penalized more heavily. At the same time, it maintains robustness across different levels of temporal specificity, effectively capturing the model’s temporal reasoning.

\paragraph{Overall Evaluation Score.}
The overall score is computed as a weighted average of the individual event, geospatial, and temporal scores. In the case of the TARA dataset, equal weights of 0.3 are assigned to the geospatial and temporal components, while the event score is weighted at 0.4. For the WikiTiLo dataset, the geospatial and temporal scores are assigned equal weights of 0.5. These weight assignments are made to prioritize the event score, as it encapsulates the primary context of the image.

\vspace*{-\baselineskip}
\begin{center}
\small
\[
\text{Overall Score} =
\begin{cases}
    0.4 \cdot ES_i + 0.3 \cdot GS_i + 0.3 \cdot TS_i \\ 
    \text{for TARA dataset} \\[1ex]
    0.5 \cdot GS_i + 0.5 \cdot TS_i \\ 
    \text{for WikiTiLo dataset}
\end{cases}
\]
\end{center}

\section{Experiments}

\paragraph{VLM Models.} For \textbf{Gemini-1.5 Pro-002}, the experiments were conducted using the batch API from Vertex AI on Google Cloud Platform (GCP).  

\noindent For \textbf{GPT-4o-mini}\textsuperscript{}\footnote{\url{https://openai.com/index/gpt-4o-mini-advancing-cost-efficient-intelligence/}}, we utilized the batch API. While OpenAI's models do not directly reveal public figures\textsuperscript{}\footnote{\url{https://openai.com/policies/usage-policies/}}, they can ground them if their identities are provided externally in the prompt. To leverage this, we used the CelebrityFaces API from AWS Rekognition\textsuperscript{}\footnote{\url{https://aws.amazon.com/rekognition/}} to identify public figures in the dataset images with a confidence score of at least 99\%. These identified public figures were then explicitly included in the prompts during the experiments to enable more precise entity recognition. For \textbf{Qwen-VL2.5-7B-Instruct}~\cite{qwen2.5-VL} ~\cite{Qwen2VL} ~\cite{Qwen-VL}, the experiments were conducted on three NVIDIA A100 Tensor Core GPUs, each with 80 GB of GPU memory, alongside 8-core CPUs with 32 GB of RAM. These resources were hosted on the Arizona State University's Sol supercomputer~\cite{HPC:ASU23}

\paragraph{Baselines.} 
We used multiple visual reasoning approaches to compare with \getreason{}. These include: reasoning-based methods such as GoodGuesser~\citep{zhang2024goodguesser} and CogBench~\citep{song2024cogbench}, which highlight limitations in spatio-temporal inference due to unstructured representations; object detection-based methodologies such as CAPTURE~\citep{dong2024benchmarking}, VIPHY~\citep{singh2023}, and EDIS~\citep{liu2023}, which focus on descriptive features and offer limited contextual reasoning; and retrieval-augmented generation (RAG) methods like Generate the Select~\citep{fu2023generate} and QR-CLIP~\citep{gao2024qrclip}, which enhance factual recall but struggle with abstract or causal reasoning. To ensure fairness, all baselines were evaluated under a unified schema using only the VLM’s internal knowledge. Additionally, we introduced two custom-designed baselines tailored to our task, Zeroshot COT and Detective Agent, whose prompts are provided in Appendix~\ref{subsec:zero} and Appendix~\ref{subsec:det}.


\paragraph{Evaluation Metric.}
To evaluate the reasoning capabilities of our framework, we compared the reasoning output generated by our model to the ground truth across three modalities on the TARA dataset on Gemini 1.5 pro-002, utilizing the Event Evaluation metric from proposed \great{} metric.

\subsection{Results and Analysis.}

Table \ref{tab:results} presents the performance comparison of our model’s three variants against 11 baseline strategies including a zeroshot Chain-of-Thoughts prompt and a Detective Agent prompt across three information types. For RAG based approaches, we modified the RAG parts of the prompts to using the pretrained knowledge base instead of internet extraction. Evaluating on the \great{} evaluation metric, we observe that across almost all cases, our framework consistently achieves the best performance for all reasoning types viz. event, temporal, and geospatial, demonstrating its superiority over other baselines. Notably, Good Guesser also performs well in geospatial reasoning, surpassing even our framework 3 out of 6 times and achieving the second highest scores in most tasks.


These results highlight the robustness of \getreason{} while also indicating that the Good Guesser baseline remains a strong contender in geospatial reasoning. 

\paragraph{Across Models Analysis} Across the three models—Gemini 1.5 Pro-002, GPT-4o-mini, and Qwen2.5VL-7B—all perform competitively in event reasoning, with Gemini leading overall. Gemini also outperforms the others in geospatial and temporal extraction. Between Qwen and GPT, Qwen consistently excels in geospatial tasks, while GPT performs better in temporal tasks. Given its larger size, Gemini shows stronger reasoning capabilities, while the smaller and similarly sized Qwen and GPT-4o-mini perform comparably, each outperforming the other in specific tasks.

\paragraph{Deduction Analysis } In order to evaluate our framework’s reasoning capabilities, we compared its output to ground truth across three modalities on the TARA dataset using Gemini 1.5 Pro-002 and the Event Evaluation metric from GREAT. \getreason{} achieves scores of 81.4\%, 76.9\%, and 70.2\% for geospatial, temporal, and event deduction, respectively. These consistently high scores across modalities indicate the model effectively extracts information from images through accurate reasoning.

\paragraph{Ablation Analysis.} We designed and implemented several alternative versions of our framework to accommodate different scenarios. Additionally, we conducted ablation studies to assess the impact of various components. We compared our approach with the Direct extraction, partial Cross extraction, where we only supply the global event information from direct extraction to the temporal and geospatial agents in the second iteration, and various variations where we remove the image, one in the prompt generation layer, the multiagentic extraction and both.

\begin{table}[h]
\centering
\small
\setlength{\tabcolsep}{2.5pt}
\begin{tabular}{llcccc}
\toprule
&  \textbf{Method} & Geo & Temp & Event & Total \\
\midrule
\multirow{5}{*}{\textbf{Tara}} & \getreason{}  & 69.4 & 38.1 & 37.7 & 47.3  \\ 
\cmidrule(lr){2-6}
& Direct Extraction & 67.4 & 33.2 & 68.6 & 57.6 \\
& *Cross Extraction & 68.2 & 35.9 & 70.3 & 59.3 \\
& -- image in ME & 44.1 & 34.4 & 68.5 & 51.2 \\
& -- image in PL + ME & 44.2 & 68.2 & 33.7 & 50.8 \\
\midrule
\multirow{3}{*}{\textbf{WikiTiLo}} & \getreason{}  & 42.4 & 34 & - &  38.2 \\ 
\cmidrule(lr){2-6}
& Direct Extraction &  41.6 & 33.8 & - & 37.7\\
& *Cross Extraction & 42.5 & 33.5 & - & 38 \\
\bottomrule
\end{tabular}
\vspace{-0.5em}
\caption{\small  Evaluation Results for Gemini 1.5 Pro-002 on TARA. \textbf{Abbreviations:} PL = prompt generator layer, ME = multiagentic extraction. WikiTiLo does not have scores for event. Here, *Cross Extraction represents Partial Cross Extraction.}
\vspace{-1.5em}
\label{tab:ablation}
\end{table}

\textit{Analysis.} Table \ref{tab:ablation} shows the results from our approach. Our findings indicate that excluding the image as an input for agents, except for the scene graph, significantly reduces performance.

\paragraph{Multi-Dimensional Error Analysis.}

We evaluate our framework across multiple reasoning benchmarks—event, temporal, and geospatial—by comparing its performance against baselines on all subsets of these benchmarks. This approach allows us to quantify the net change in error between our model and the baselines across individual and combined benchmarks.

\begin{table}[h]
\setlength{\tabcolsep}{2.3pt}
\small
    \centering
        \begin{tabular}{l c c c c c c c}
        \toprule
          \textbf{Baseline}   & TS & GS & ES & TS & TS & GS & All \\
                  &  & & & + GS &  + ES &  + ES & Three \\  
          \midrule
             COT$_{\text{zeroshot}}$ & 5.7 & 15.9 & 36.8 & 3.9 & 9.4 & 14.2 & 3.5 \\
DA Prompt & 15.7 & 19.6 & 31.0 & 7.4 & 14.4 & 16.6 & 6.1 \\
VIPHY & 35.4 & 37.1 & 64.4 & 18.2 & 33.5 & 35.6 & 16.1 \\
CogBench & 40.3 & 30.5 & 60.5 & 18.3 & 36.7 & 30.4 & 16.0 \\
CAPTURE & 25.4 & 36.2 & 63.9 & 13.8 & 25.3 & 35.1 & 12.3 \\
GTS & 25.1 & 19.8 & 51.9 & 9.3 & 23.4 & 20.2 & 7.9 \\
QR-CLIP & 27.4 & 20.9 & 47.1 & 10.3 & 23.7 & 20.8 & 8.6 \\
EDIS & 35.5 & 36.4 & 59.0 & 17.7 & 31.5 & 33.7 & 15.2 \\
Good Guesser & 46.8 & -15.1 & 51.7 & 3.2 & 40.9 & 0.9 & 3.6 \\
            \bottomrule
        \end{tabular}
    \vspace{-0.5em}
    \caption{ \small Net percentage gains where \getreason{} outperforms individual baselines for TARA with Gemini 1.5 Pro-002. GTS = Generate then Select. DA = Detective Agent}
    \label{tab:error_analysis}
    \vspace{-1.5em}
\end{table}

We compute the frequency \( f_1 \) of samples where our framework outperforms the baselines, indicating error reduction, and the frequency \( f_2 \) where baselines outperform our framework, indicating error increase. The net error change \( NC_i \) for a given combination \( i \) is defined as:  

\vspace{-0.5em}
\begin{equation} \label{eq:net_error_change}
\small
NC_i = \frac{(f_1 - f_2)}{N} \times 100
\end{equation}

where \( N \) denotes the total number of samples. Table~\ref{tab:error_analysis} presents the computed net error changes across different benchmark combinations.

\subsection{\getreason{} Effectiveness} 

There are two primary reasons \getreason{} consistently outperforms both existing and internally developed baselines:

\paragraph{(1) Methodological Advances.} Our approach introduces several core innovations:  
\textit{(a) Hierarchical Multi-Agent Design}: By decomposing the task across specialized agents, our method reduces VLM overload, resets context between subtasks, and mitigates hallucinations—unlike monolithic prompting schemes.  
\textit{(b) Structured Prompting}: We employ role-specific prompts with controlled information flow, yielding more consistent and interpretable outputs compared to ad-hoc prompting strategies.  
\textit{(c) Schema-Guided Reasoning}: Our framework blends schema extraction with flexible reasoning, enabling precision without sacrificing adaptability—addressing common limitations of unstructured methods.  
\textit{(d) Systematic Cue Integration}: Temporal, spatial, and event-specific cues are explicitly handled, unlike prior models that overlook or inconsistently process such signals (see Section~\ref{sec:baselines-composition}).  
\textit{(e) Scalable Evaluation}: \getreason{} is evaluated on over 17,000 public event images, thus a more reliable assessment than prior work typically benchmarked on $\sim$2,000 samples.

\paragraph{(2) Baseline Limitations.}  Competing methods—spanning object-centric captioning, shallow reasoning, and retrieval-augmented generation—struggle with contextual grounding, prompt coherence, and resilience to misinformation. Our structured, agent-based design addresses these issues holistically.

\section{Comparison with Related Work}
\label{sec:baselines-composition}

\paragraph{Reasoning-Based Approaches.}  
~\citet{zhang2024goodguesser} introduced GoodGuesser, a benchmark for assessing VLMs' spatio-temporal reasoning through VQA tasks, revealing limitations in contextual understanding. Similarly, ~\citet{song2024cogbench} proposed CogBench, evaluating LVLMs across eight domains to quantify reasoning gaps. However, both rely on unstructured approaches, making event extraction challenging due to high context retention demands.

\paragraph{Object Detection-Based Approaches.}  

Recent works in object detection, such as ~\citet{dong2024benchmarking} propose CAPTURE for fine-grained image captioning, ~\citet{singh2023} develop VIPHY to assess physical commonsense reasoning, and ~\citet{liu2023} introduce EDIS for entity-driven retrieval. However, these methods focus on descriptive features rather than contextual reasoning.

\paragraph{RAG-Based Approaches.}  

Recent approaches, such as Generate the Select~\citet{fu2023generate} and QR-CLIP~\citet{gao2024qrclip}, have advanced the integration of external knowledge into visual models. Generate the Select focuses on evaluating world knowledge in VQA tasks, emphasizing factual recall but lacking deeper contextual reasoning. QR-CLIP enhances CLIP-based retrieval through query reformulation, improving location and time reasoning but not structured inference. While both methods improve factual accuracy, they fall short in supporting complex reasoning processes, such as abstract or causal inference.

\section{Conclusion}
A structured multi-agent extraction approach effectively infers event context, temporal information, and geospatial location while ensuring robustness against component failures and hallucinations. By assigning specific tasks to individual agents, this method maintains a controlled reasoning process, systematically integrating relevant contextual information. This structured approach leverages the subjective nature of reasoning while constraining factual information within well-defined tasks to enhance accuracy. Furthermore, we introduce a metric that evaluates both subjective and objective information, ensuring that the scoring process aligns with the underlying reasoning framework. 

\section*{Limitations}
The main limitation of this approach is the language barrier, as it relies on English-language datasets. Additionally, the TARA dataset’s extraction method results in many images being unrelated to their corresponding articles. Our analysis is based on a subset of TARA, whereas a more comprehensive evaluation could be conducted on the full 61k dataset or even larger alternatives like SHERLOCK~\cite*{hessel2022abduction} and BreakingNews dataset~\cite*{ramisa-etal-2017-breakingnews} along with the comparison with newer studies like PuzzleGPT~\cite*{ayyubi2025puzzlegpt} to better assess data loss and the study’s limitations. Furthermore, as context extraction depends on model reasoning and Vision-Language Models (VLMs) are rapidly advancing, our findings may not fully reflect the latest developments. Although we used three competitive models, more robust reasoning based alternatives like OpenAI-o1\textsuperscript{}\footnote{\url{https://openai.com/o1/}}
, OpenAI-o3-mini\textsuperscript{}\footnote{\url{https://openai.com/index/openai-o3-mini/}} and DeepSeek-R1~\cite{deepseek2025r1} have since emerged.

\section*{Ethics}
Our study utilizes the existing TARA and WikiTiLo datasets, thereby inheriting any pre-existing biases and constraints within them; however, it does not introduce any additional biases or constraints. During dataset augmentation, we ensured that no extraneous information was incorporated by exclusively using the existing metadata and labels. All hosted models and pre-trained weights were employed strictly for research purposes, and our study remains confined to academic research. As our work involves facial recognition, we have limited its application to images sourced solely from these publicly available datasets. Additionally, we have utilized the Celebrity Facial Recognition API provided by AWS while strictly adhering to its privacy policies and guidelines.

\section*{Acknowledgements}
We thank the anonymous reviewers for their helpful feedback. We also thank Zenia Shaikh for her valuable assistance and contributions to this work.  We thank the Complex Data Analysis and Reasoning Lab at Arizona State University for computational support, and Krishna Singh Rajput for his support during the submission process. Lastly, thanks to our lab cat, Coco, for keeping our professor company and adding some much-needed humor during deadline crunches.

\bibliography{acl_latex}

\appendix

\label{apd}
\section{Individual Agentic Prompts:}

\subsection{Scene Graph Agent prompt.}
\label{subsec:sgprompt}

\begin{lstlisting}
You are an expert in analyzing and understanding visual images to generate detailed and structured scene graphs in JSON-formatted outputs. You perform the following two tasks sequentially:

Scene Graph Generation: For every image provided, identify all entities, objects, attributes, and relationships in the scene. The graph must include all common entities, whether singular or plural, such as "car," "person," "crowd," "tree," "building," etc., as well as their properties and the spatial or contextual relationships among them. Your goal is to create a complete and accurate representation of all visible elements in the scene. 

Graph Enhancement: Refine the scene graph by identifying and replacing the generic descriptions of entities that correspond to public figures provided in the input list. For instance, if a "person" in the scene is identified as a specific public figure, replace "person" with their provided name. During this process, ensure that:
All entities, whether identified or unidentified, remain in the graph. 

All relationships between objects and entities are preserved.

Unidentified or unrecognized entities are left unchanged but still included in the graph.

Your responses must be concise but thorough, breaking down the image into its components and interactions while ensuring that the scene graph remains fully intact, comprehensive, and user-friendly. Do not prune out entities or relationships during any part of the process.

The objective is to provide users with a refined scene graph that includes all visible entities and incorporates specific public figure names where applicable, enhancing both clarity and usability.

You may or may not be provided with the public figures present in the image. If present, ground them to their common nouns in their entities.

You will return the output in the required response format. It is absolutely imperative that you return the JSON output. You will extract the information required for the response format. If no plausible guess can be made for a field, output NA.
\end{lstlisting}

\subsection{Abstract Agent prompt.}
\label{aa_prompt}

\begin{lstlisting}
You are a highly advanced AI agent designed to analyze abstract relationships between entities in visual scenes and generate a comprehensive abstract idea behind the image.

You will be given an Image (e.g., URL or binary data).

You will also be given a Scene Graph in JSON format that details the objects, attributes, and relationships within the image.

Your task is to:

Derive an abstract\textunderscore idea that captures the overall essence or message of the scene.

Provide your reasoning that concisely explains how you arrived at that abstract idea based on the provided Scene Graph.

Important:

You may reason abstractly or hypothetically about the relationships and significance of objects in the scene.

Your response must be concise yet detailed, focusing on explaining the underlying connections and themes in the image.

You will return the output in the required response format. It is absolutely imperative that you return the JSON output. You will extract the information required for the response format. If no plausible guess can be made for a field, output NA.

Scene Graph: <generated scene graph>

\end{lstlisting}
\subsection{Prompt Agent Prompt.}\label{pa_prompt} 

\begin{lstlisting}

You are a prompt generating agent designed to assist in analyzing images and their associated scene graphs by formulating tailored queries or prompts for specified agents. Your primary goal is to provide specific, context-relevant prompts for each agent, enabling them to deliver detailed and role-specific insights about the image. You should understand the roles of the agents and generate clear, concise, and actionable prompts tailored to their individual expertise and capabilities and the reasoning behind the specific prompts. You adapt to diverse scenarios while maintaining precision and relevance in the prompts.

Agents to use and descriptions - 

Global event specialist - Analyzes global events using multimodal inputs for detailed reasoning. 

Temporal specialist - Analyzes images to deduce time of day (for eg day, night, etc) or period (for eg date, year, decade, century, etc).

Geospatial specialist - Analyzes locations from images and scene graphs with detailed reasoning. You will return the output in the required response format. It is absolutely imperative that you return the JSON output. You will extract the information required for the response format. If no plausible guess can be made for a field, output NA.
Scene Graph:<abstract added scene graph>
\end{lstlisting}

\subsection{Geospatial Agent Direct Extraction prompt.}
\label{ga_dir_prompt} 

\begin{lstlisting}
You are a custom agent designed to analyze an image and any accompanying contextual information to deduce geospatial details about the scene depicted. The details to determine are:
- Country
- State_or_Province
- City
Using the visual elements of the image and the provided contextual information, you must determine the best possible value for each of these fields. If you cannot deduce a particular field with reasonable confidence, output "NA" for that field.

Rules:
- You must provide exactly one value for each of the following fields: "Country," "State_or_Province," and "City." You will return the output in the required response format. It is absolutely imperative that you return the JSON output.You will extract the information required for the response format. If no plausible guess can be made for a field, output NA.",

Scene Graph: <abstract added scene graph>
Prompt: <prompt agent generated geospatial agent prompt>
\end{lstlisting}

\subsection{Temporal Agent Direct Extraction prompt.}\label{ta_dir_prompt}
\begin{lstlisting}
You are an agent designed to deduce the time of day, time period, or year depicted in an image based on its visual elements, the scene graph, and a descriptive prompt. 

When analyzing an image, consider the following: 

- Time of Day: Use clues like lighting, shadows, and visible celestial objects (e.g., sun, moon, stars).

- Time Period/Year: Infer the era or approximate year based on historical or cultural details such as clothing, architecture, technology, or objects depicted.

- Best Guess: If conclusive reasoning is not possible, provide the most plausible estimate and explain your reasoning. You will return the output in the required response format. It is absolutely imperative that you return the JSON output. You will extract the information required for the response format. If no plausible guess can be made for a field, output NA.

Scene Graph: <abstract added scene graph>
Prompt: <prompt agent generated temporal agent prompt>    
\end{lstlisting}

\subsection{Event Agent Direct Extraction prompt.}\label{ea_dir_prompt} 

\begin{lstlisting}

You are an agent designed to analyze global events using multimodal inputs. You combine the given inputs to provide a detailed, concise analysis of the primary event asssociated with this image along with the background behind the primary event. Use logical reasoning and general world knowledge to address gaps in the inputs. Output in the following structure:  

1. Global Event Context: Provide a brief summary of the event depicted in the image. 

2. Analysis: Offer detailed reasoning combining the provided structured contextual clues to explain the event and its implications.  

3. Best Guess: Make logical assumptions if the inputs are inconclusive, ensuring relevance to the global context. Tailor responses to be clear, neutral, and informative, avoiding bias or unfounded speculation.You will return the output in the required response format. It is absolutely imperative that you return the JSON output. You will extract the information required for the response format. If no plausible guess can be made for a field, output NA.

Scene Graph: <abstract added scene graph>
Prompt: <prompt agent generated event agent prompt>

\end{lstlisting}

\subsection{Geospatial Agent Cross Extraction prompt.}\label{ga_cr_prompt}
\begin{lstlisting}
You are a custom agent designed to analyze an image and any accompanying contextual information to deduce geospatial details about the scene depicted, paying special attention to the added event and temporal information along with the event's background. It is important to note that any null, NA, unknown values in the provided are ignored. The details to determine are:
  - Country
  - State_or_Province
  - City

Using the visual elements of the image and the provided contextual information, you must determine the best possible value for each of these fields. If you cannot deduce a particular field with reasonable confidence, output "NA" for that field.

Rules:
  - You must provide exactly one value for each of the following fields: "Country," "State_or_Province," and "City". You will return the output in the required response format. It is absolutely imperative that you return the JSON output. You will extract the information required for the response format. If no plausible guess can be made for a field, output NA.

Scene Graph: <abstract added scene graph + event agent output + temporal agent output>
Prompt: <prompt agent generated geospatial agent prompt>
\end{lstlisting}

\subsection{Temporal Agent Cross Extraction prompt.}\label{ta_cr_prompt}

\begin{lstlisting}
You are an agent designed to deduce the time of day, time period, or year depicted in an image based on its visual elements, the scene graph, and a descriptive prompt paying special attention to the added event and geospatial information along with the event's background. It is important to note that any null, NA, unknown values in the provided are ignored. When analyzing an image, consider the following: 

- Time of Day: Use clues like lighting, shadows, and visible celestial objects (e.g., sun, moon, stars).

- Time Period/Year: Infer the era or approximate year based on historical or cultural details such as clothing, architecture, technology, or objects depicted.

- Best Guess: If conclusive reasoning is not possible, provide the most plausible estimate and explain your reasoning. You will return the output in the required response format. It is absolutely imperative that you return the JSON output.You will extract the information required for the response format. If no plausible guess can be made for a field, output NA.
Scene Graph: <abstract added scene graph + event agent output + geospatial agent output>
Prompt: <prompt agent generated temporal agent prompt>
\end{lstlisting}

\subsection{Event Agent Cross Extraction prompt.}\label{ea_cr_prompt} 
\begin{lstlisting}
You are an agent designed to analyze global events using multimodal inputs. You combine the given inputs, paying special attention to the added temporal and spatial information to provide a detailed, concise analysis of the primary event associated with this image along with the background behind the primary event.

It is important to note that any null, NA, or unknown values in the provided temporal and spatial information are ignored. Use logical reasoning and general world knowledge to address gaps in the inputs.

Output in the following structure:  
1. Global Event Context: Provide a brief summary of the event depicted in the image.  
2. Analysis: Offer detailed reasoning combining the provided structured contextual clues to explain the event and its implications.  
3. Best Guess: Make logical assumptions if the inputs are inconclusive, ensuring relevance to the global context.

Tailor responses to be clear, neutral, and informative, avoiding bias or unfounded speculation. You will return the output in the required response format. It is absolutely imperative that you return the JSON output. You will extract the information required for the response format. If no plausible guess can be made for a field, output NA.

Scene Graph: <abstract added scene graph + geospatial agent output + temporal agent output>  
Prompt: <prompt agent generated event agent prompt>

\end{lstlisting}

\subsection{Zeroshot COT prompt.}
\label{subsec:zero}
\begin{lstlisting}
Analyse this image and get the following information about this image:

temporal information:
- century
- decade
- year
- month
- day

geospatial information:
- country
- state_or_province
- city

event: the event in this image  
background: the most relevant background related to the event

Let's think step-by-step. You may or may not be provided with the public figures present in the image. If present, ground them to their common nouns in their entities. You will return the output in the required response format. It is absolutely imperative that you return the JSON output. You will extract the information required for the response format. If no plausible guess can be made for a field, output NA.

\end{lstlisting}

\subsection{Detective Agent prompt.}
\label{subsec:det}
\label{detective_prompts}

\begin{lstlisting}

You are an expert detective. You specialize in analyzing a scene and reasoning beyond the scene. For a given image, you will look at the scene and all the probable inferences and reason based on them to reason beyond the image to retrieve your best guess for the event that is going on or has happened in this image, the background related to this image. The geospatial information and the temporal information.You may or may not be provided with the public figures present in the image. If present, ground them to their common nouns in their entities.You will return the output in the required response format. It is absolutely imperative that you return the JSON output. You will extract the information required for the response format. If no plausible guess can be made for a field, output NA.

\end{lstlisting}

\section{Working Examples}
\label{sec:working_section}

For the image \ref{fig:indonesian_president}, the image is input into the Scene Graph Agent with the Scene Graph Agent prompt to output the basic Scene Graph: 

\begin{lstlisting}
{
  "id": 317,
  "response": {
    "entities": [
      {
        "attributes": [
          { "attribute": "wearing", "value": "suit" }
        ],
        "entity": "Joko Widodo"
      },
      {
        "attributes": [
          { "attribute": "wearing", "value": "robes" }
        ],
        "entity": "person"
      },
      {
        "attributes": [
          { "attribute": "wearing", "value": "suit" }
        ],
        "entity": "person"
      },
      {
        "attributes": [
          { "attribute": "wearing", "value": "hijab" }
        ],
        "entity": "person"
      },
      {
        "attributes": [
          { "attribute": "wearing", "value": "suit" }
        ],
        "entity": "person"
      },
      {
        "attributes": [
          { "attribute": "wearing", "value": "suit" }
        ],
        "entity": "person"
      },
      {
        "attributes": [
          { "attribute": "wearing", "value": "suit" }
        ],
        "entity": "person"
      },
      {
        "attributes": [
          { "attribute": "wearing", "value": "suit" }
        ],
        "entity": "person"
      },
      {
        "attributes": [
          { "attribute": "wearing", "value": "suit" }
        ],
        "entity": "person"
      },
      {
        "attributes": [
          { "attribute": "type", "value": "religious" }
        ],
        "entity": "book"
      },
      { "attributes": [], "entity": "flowers" },
      { "attributes": [], "entity": "microphones" },
      { "attributes": [], "entity": "podium" }
    ],
    "relationships": [
      {
        "reasoning": "Joko Widodo is holding the book",
        "relationship": "holding"
      },
      {
        "reasoning": "Joko Widodo is standing at the podium",
        "relationship": "at"
      },
      {
        "reasoning": "The microphones are on the podium",
        "relationship": "on"
      },
      {
        "reasoning": "The flowers are on the podium",
        "relationship": "on"
      },
      {
        "reasoning": "The person is holding up the book for Joko Widodo",
        "relationship": "holding for"
      }
    ]
  }
}
\end{lstlisting}

This Basic scene graph, along with the Abstract Agent prompt and the image are input to the Abstract Agent to output: 

\begin{lstlisting}
{
    "id": 317,
    "response": {
        "abstract_idea": {
            "idea": "A ceremonial oath or affirmation.",
            "reasoning": "Joko Widodo, wearing a suit, is holding a book, likely a religious text, while another person in robes holds it up for him.  This, combined with the formal attire of the other individuals in suits, the presence of microphones and a podium adorned with flowers, suggests a formal ceremony, possibly an inauguration or oath of office.  The act of holding the book symbolizes a commitment to uphold principles and responsibilities associated with the position."
        }
    }
}
\end{lstlisting}

the abstract\textunderscore idea from this is added to the scene\textunderscore graph to create:

\begin{lstlisting}
    {
    "entities": [
        {
            "attributes": [
                {
                    "attribute": "wearing",
                    "value": "suit"
                }
            ],
            "entity": "Joko Widodo"
        },
        {
            "attributes": [
                {
                    "attribute": "wearing",
                    "value": "robes"
                }
            ],
            "entity": "person"
        },
        {
            "attributes": [
                {
                    "attribute": "wearing",
                    "value": "suit"
                }
            ],
            "entity": "person"
        },
        {
            "attributes": [
                {
                    "attribute": "wearing",
                    "value": "hijab"
                }
            ],
            "entity": "person"
        },
        {
            "attributes": [
                {
                    "attribute": "wearing",
                    "value": "suit"
                }
            ],
            "entity": "person"
        },
        {
            "attributes": [
                {
                    "attribute": "wearing",
                    "value": "suit"
                }
            ],
            "entity": "person"
        },
        {
            "attributes": [
                {
                    "attribute": "wearing",
                    "value": "suit"
                }
            ],
            "entity": "person"
        },
        {
            "attributes": [
                {
                    "attribute": "wearing",
                    "value": "suit"
                }
            ],
            "entity": "person"
        },
        {
            "attributes": [
                {
                    "attribute": "wearing",
                    "value": "suit"
                }
            ],
            "entity": "person"
        },
        {
            "attributes": [
                {
                    "attribute": "type",
                    "value": "religious"
                }
            ],
            "entity": "book"
        },
        {
            "attributes": [],
            "entity": "flowers"
        },
        {
            "attributes": [],
            "entity": "microphones"
        },
        {
            "attributes": [],
            "entity": "podium"
        }
    ],
    "relationships": [
        {
            "reasoning": "Joko Widodo is holding the book",
            "relationship": "holding"
        },
        {
            "reasoning": "Joko Widodo is standing at the podium",
            "relationship": "at"
        },
        {
            "reasoning": "The microphones are on the podium",
            "relationship": "on"
        },
        {
            "reasoning": "The flowers are on the podium",
            "relationship": "on"
        },
        {
            "reasoning": "The person is holding up the book for Joko Widodo",
            "relationship": "holding for"
        }
    ],
    "abstract_idea": "A ceremonial oath or affirmation."
}
\end{lstlisting}

This augmented scene\textunderscore graph along with the Prompt Agent's prompt and the image is input into the Prompt Agent to output:

\begin{lstlisting}
 {
    "id": 317,
    "response": {
        "global_event_specialist": {
            "prompt": "Based on the image and scene graph information, analyze the event depicted. Consider the individuals present, their attire, the objects visible (such as the book and podium), and the overall setting. Provide insights into the nature and significance of the ceremony, including its potential cultural or political context. Is this a globally recognized event?",
            "reasoning": "This prompt guides the global event specialist to analyze the event by considering visual cues like attire and objects, combined with the scene graph's relational information. It encourages a deep understanding of the ceremony's significance and global relevance."
        },
        "spatial_specialist": {
            "prompt": "Analyze the image and scene graph to determine the location of the event. Consider the architectural style, furniture, and any other environmental clues. Provide specific details about the venue if possible. Is the event taking place indoors or outdoors? What can you infer about the importance of the location?",
            "reasoning": "This prompt directs the location specialist to focus on spatial reasoning. It asks them to identify the venue, environment and significance of the location using clues from the image and scene graph"
        },
        "temporal_specialist": {
            "prompt": "Based on the image and scene graph, determine the time of day or period when this event likely took place. Consider factors such as lighting, attire, and the nature of the event. Provide justification for your deduction. Can you estimate the decade, or year?",
            "reasoning": "This prompt instructs the time specialist to deduce the timeframe of the event by analyzing visual cues such as lighting, attire, and context from the scene graph's abstract idea. This prompt encourages deeper temporal analysis and reasoning."
        }
    }
}   
\end{lstlisting}

Using this, the augmented scene\textunderscore graph along with the Temporal Agent Direct Extraction prompt and image along with the custom prompt:
"Based on the image and scene graph, determine the time of day or period when this event likely took place. Consider factors such as lighting, attire, and the nature of the event. Provide justification for your deduction. Can you estimate the decade, or year?"
is fed to the Temporal Agent for Direct Extraction, similarly, "Based on the image and scene graph information, analyze the event depicted. Consider the individuals present, their attire, the objects visible (such as the book and podium), and the overall setting. Provide insights into the nature and significance of the ceremony, including its potential cultural or political context. Is this a globally recognized event?" for the Event Agent and "Analyze the image and scene graph to determine the location of the event. Consider the architectural style, furniture, and any other environmental clues. Provide specific details about the venue if possible. Is the event taking place indoors or outdoors? What can you infer about the importance of the location?" for the Geospatial Agent are fed respectively to output:

Temporal Agent:

\begin{lstlisting}
{
    "id": 317,
    "response": {
        "century": "21st",
        "day": "NA",
        "decade": "2010s,2020s",
        "month": "NA",
        "reasoning": "The image shows Joko Widodo, the president of Indonesia, taking an oath. His first term began in 2014 and his second in 2019, hence the 2010s and 2020s. The scene suggests a formal indoor event, but there's no indication of the specific time of day, month, or date.",
        "time_of_day": "NA",
        "year": "NA"
    }
}
\end{lstlisting}

Geospatial Agent: 

\begin{lstlisting}
{
    "id": 317,
    "response": {
        "city": "Jakarta",
        "country": "Indonesia",
        "reasoning": "The scene graph identifies Joko Widodo, the President of Indonesia.  Presidential inaugurations in Indonesia take place in Jakarta, the capital city.",
        "state_or_province": "NA"
    }
}
\end{lstlisting}

Event Agent: 

\begin{lstlisting}
    {
    "id": 317,
    "response": {
        "background": {
            "reasoning": "The image shows Joko Widodo, the President of Indonesia, taking an oath of office. He is dressed in a formal suit and holding a religious text, likely the Quran, which is customary in Indonesian presidential inaugurations. The setting appears to be a formal government building, likely the People's Consultative Assembly (MPR) building in Jakarta, where presidential inaugurations take place.  The other individuals present are likely government officials and dignitaries, dressed in formal attire. The flowers and ceremonial aspects further suggest a significant state event.",
            "value": "Joko Widodo's second term inauguration as President of Indonesia in 2019."
        },
        "event": {
            "reasoning": "The individual being sworn in is Joko Widodo, recognizable from publicly available images.  The setting, the presence of other formally dressed individuals, and the ceremonial nature of the event strongly suggest a presidential inauguration.  The religious book being held is consistent with Indonesian tradition, where the President takes an oath on a religious text. The event is consistent with other reports and imagery of Widodo's inaugurations.",
            "value": "Presidential Inauguration"
        }
    }
}
\end{lstlisting}

These results are cross fed according to the method described in the \ref{fig:GETReason_architecture} along with the respective Cross Extraction prompts, the augmented scene\textunderscore graph, the image and the respective custom prompts, for the Cross extraction, to finally generate:

Temporal Agent: 

\begin{lstlisting}
    {
    "id": 317,
    "response": {
        "century": "21st",
        "day": "20th",
        "decade": "2010s",
        "month": "October",
        "reasoning": "The event is Joko Widodo's second term presidential inauguration, which took place on October 20, 2019. The image shows an indoor setting with artificial lighting, suggesting a daytime event.  The background information from the event and scene graph confirms the date.",
        "time_of_day": "Day",
        "year": "2019"
    }
}
\end{lstlisting}

Geospatial Agent: 

\begin{lstlisting}
    {
    "id": 317,
    "response": {
        "city": "Jakarta",
        "country": "Indonesia",
        "reasoning": "The event is Joko Widodo's second term inauguration as President of Indonesia. Presidential inaugurations in Indonesia take place at the People's Consultative Assembly (MPR) building in Jakarta.",
        "state_or_province": "NA"
    }
}
\end{lstlisting}

Event Agent: 

\begin{lstlisting}
    {
    "id": 317,
    "response": {
        "background": {
            "reasoning": "Joko Widodo is being sworn in for his second term as President of Indonesia.  He is taking the oath of office on a Quran, signifying his commitment to uphold the principles of the Indonesian constitution and the values of the nation.  This ceremony marks the beginning of his next five-year term, during which he will lead the country and implement his policy agenda.  Indonesia is the world's most populous Muslim-majority country, and the presidential inauguration is a significant event in the nation's political life.  The presence of high-ranking officials and dignitaries signifies the importance of this democratic transition of power.",
            "value": "Joko Widodo's second term presidential inauguration in Indonesia."
        },
        "event": {
            "reasoning": "The image depicts Joko Widodo taking the oath of office during his second presidential inauguration. The presence of a religious book (Quran), his formal attire, and the ceremonial setting strongly suggest an inauguration ceremony. The individuals surrounding him are likely high-ranking officials and dignitaries.  Given his attire and the ceremonial context, the central event is the inauguration itself.",
            "value": "Presidential Inauguration"
        }
    }
}
\end{lstlisting}

\section{Robustness of GETReason.}

Our framework relies on multiple VLM-generated outputs, each carrying a risk of hallucinations. As the number of VLMs increases, so does the potential for erroneous or misleading information, which can affect output reliability. While our structured approach mitigates hallucinations to some extent, it does not eliminate them entirely. 

\begin{table}[htbp]
\centering
\small
\begin{tabular}{l l cccc}
\toprule
\textbf{Agent Removed} & Geo & Temp & Event & Total \\
\midrule
- abstract generator & 69.2 & 38.9 & 71.3 & 61 \\
- prompt generator & 69.1 & 39.0 & 71 & 60.7 \\
- scene-graph agent & 69.0 & 38.2 & 71.2 & 60.6 \\
\bottomrule
\end{tabular}
\vspace{-0.5em}
\caption{ \small Evaluation of model robustness on TARA dataset on Gemini 1.5 Pro-002.}
\label{tab:robustness}
\vspace{-1.5em}
\end{table}

\textit{Analysis.} To assess their impact, we analyze performance with certain components removed as seen in Table~\ref{tab:robustness}. The results indicate that reducing components—thus decreasing hallucination-prone generations—does not significantly affect overall scores. This suggests that our framework remains robust to noise introduced by hallucinations and agent failures.

\subsection{Additional Ablation Results}

Tables \ref{tab:qwen_results} and \ref{tab:gpt_results} show additional results of direct and partial cross extract of \getreason{} on QwenVL2.5-7B-Instruct and GPT-40 mini, achieved on TARA and WikiTiLo.

\begin{table}[htbp]
\small
\centering
\setlength{\tabcolsep}{6.5pt}
\begin{tabular}{l l cccc}
\toprule
 & & \multicolumn{4}{c}{\textbf{QwenVL2.5-7B-Instruct}} \\
\cmidrule(lr){3-6}
\textbf{Dataset} & \textbf{Method} & Geo & Temp & Event & Total \\
\midrule
\multirow{2}{*}{TARA} & DE & 49.8 & 17.8 & 63.1 & 45.5 \\ 
& *CE & 56.8 & 22.7 & 63.2 & 49.1 \\
\midrule
\multirow{2}{*}{WikiTiLo} & DE & 43.2 & 23.7 & - & 33.5 \\
& *CE & 33.4 & 27.9 & - & 22.3 \\
\bottomrule
\end{tabular}
\vspace{-0.75em}
\caption{\small Evaluation Results (\%) for QwenVL2.5-7B-Instruct. DE = Direct Extraction, *CE = Cross Extraction.}
\label{tab:qwen_results}
\end{table}

\begin{table}[h]
\small
\centering
\setlength{\tabcolsep}{6.5pt}
\begin{tabular}{l l cccc}
\toprule
 & & \multicolumn{4}{c}{\textbf{GPT-4o mini}} \\
\cmidrule(lr){3-6}
\textbf{Dataset} & \textbf{Method} & Geo & Temp & Event & Total \\
\midrule
\multirow{2}{*}{TARA} & DE & 44.2 & 42.1 & 66.8 & 52.6 \\ 
& *CE & 43.6 & 44.6 & 68.5 & 53.8 \\
\midrule
\multirow{2}{*}{WikiTiLo} & DE & 17.29 & 26.2 & - & 21.7 \\
& *CE & 17.1 & 26.8 & - & 21.9 \\
\bottomrule
\end{tabular}
\vspace{-0.75em}
\caption{\small Evaluation Results (\%) for GPT-4o mini. DE = Direct Extraction, *CE = Cross Extraction.}
\label{tab:gpt_results}
\vspace{-2.0em}
\end{table}

\section{Further Discussion} 
In this section we discussion error propagation in \getreason{}, RAG vs \getreason{}, and usability and scalability of \getreason{}. 

\subsection{Error propagation}
Our architecture inherently mitigates error propagation through its structured design. Each stage extracts structured JSON outputs, beginning with the scene graph. By explicitly instructing agents to generate the necessary information for the scene graph, we maintain a deterministic structure while allowing free-form values to capture subjective entities, attributes, and relationships. This approach significantly reduces the likelihood of hallucinations and errors.

Even when errors occur due to mis-generated values, subsequent agents mitigate their impact by referencing the original image provided as input alongside previously generated outputs. This prevents cascading biases from earlier stages, enhancing robustness and preventing catastrophic error propagation. Additionally, our prompt generation mechanism reinforces the focus on correct visual cues by maintaining a direct reference to the image, further improving reliability by maintaining focus on the essential aspects of the image necessary for the deduction tasks.

To evaluate this very aspect, we simulated agent failures by removing specific agents (see Table~\ref{tab:robustness}). The approach remained stable, as other agents, still provided with the image, maintained performance. However, ablation studies (see Table~\ref{tab:ablation}) show that when the image is removed from subsequent agents, performance drops significantly. This confirms that excluding the image increases the likelihood of error propagation and hallucinations.

Regarding our multi-agent framework, we recognize the need for further clarification and will revise the relevant section accordingly. Each agent generates structured JSON outputs:

\begin{itemize}[left=0pt]
    \item \textbf{Temporal agent}: \{century, decade, year, month, date, reasoning\}
    \item \textbf{Geospatial agent}: \{country, state\_or\_province, city, reasoning\}
    \item \textbf{Event agent}: \{event, event\_reasoning, background, background\_reasoning\}
\end{itemize}

Similar to the scene graph stage, this structured approach inherently reduces initial errors. Additionally, we employ a cross-extraction strategy, where each agent receives contextual inputs from all relevant sources---the scene graph, prompts, and outputs from the other two agents---to further solidify the accuracy of the deductions. Our ablation studies confirm that cross-extraction outperforms direct and partial cross-extraction, reinforcing its role in minimizing error propagation.

\subsection{RAG vs GETReason}
The key distinction between our approach and Retrieval-Augmented Generation (RAG) lies in the fundamental nature of the problem statement. Our method is designed to reason about temporal, geospatial, and event-related information purely from the visual cues in an image, rather than retrieving it from external sources such as the internet.

While retrieval-based methods can extract information from online databases, they face significant challenges, including unreliable sources, misinformation, incomplete views, distributed information, and API limitations, as discussed by ~\cite{zeng2025worse}. In contrast, our approach remains independent of external knowledge and focuses entirely on visual reasoning to infer meaningful information.

Implementing a retrieval-based solution for this problem is straightforward---for instance, a reverse image search using Google's RIS engine could extract image-related metadata via APIs, as discussed by ~\cite{tonglet2025cove, tonglet2024image}. However, such an approach fundamentally bypasses the reasoning process by looking for the correct answer at an external source rather than analyzing and interpreting the image itself. As a result, RAG-based methods fail to capture the deeper contextual understanding required for applications like journalism and sentiment analysis, where the interpretation of an image is crucial.

Moreover, a retrieval-based approach transforms the problem into an engineering challenge rather than a research problem. By functioning primarily as a lookup mechanism with LLMs as a tool for retrieval, it does not contribute to advancing the field of visual reasoning and multimodal understanding.



\subsection{Usability and Scalability of GETReason}
GETReason’s multi-agent approach is computationally expensive, requiring multiple large model passes per image. While the accuracy improvements are substantial, the scalability of this system for real-world deployment (e.g., in a high-traffic news setting) is unclear. Comparing inference speed with simpler approaches (such as directly prompting a large VLM) would provide useful insight.

The computational efficiency of our approach becomes particularly relevant when applied in real-world scenarios. Use cases such as journalism and fact-checking prioritize rich contextual understanding over raw speed. However, in high-traffic environments, such as real-time news processing, our approach can be parallelized across all data points to enhance scalability.

While our agents are sequentially dependent, their computations can be batched---allowing all data points for a single agent to be processed simultaneously. This significantly reduces the average processing time per data point.

Moreover, the three agents in the multi-agentic framework operate independently, relying only on the input prompt and prior outputs, which enables parallel execution. The same applies to cross-extraction, further optimizing efficiency.

In fact, we used both these optimizations for running our experiments, which allowed us to evaluate 9 baselines alongside our approach on 3 different models with a total dataset size of about 17{,}000 samples.

By leveraging these two levels of parallelization, our approach effectively balances processing speed and output quality, ensuring scalability without compromising accuracy.

\end{document}